\documentclass{sig-alternate}
\usepackage[nocapsections]{fixacm}
\usepackage[bookmarks=false,pdfdisplaydoctitle=true]{hyperref}

\usepackage{fancyhdr}
\usepackage{epstopdf}
\DeclareFixedFont{\auacc}{OT1}{phv}{m}{n}{12} 
\DeclareFixedFont{\afacc}{OT1}{phv}{m}{n}{10} 
\makeatletter
\let\@copyrightspace\relax
\makeatother

\begin{document}

\title{Joint Belief and Intent Prediction for Collision Avoidance in Autonomous Vehicles}
%
%
%
%
%

\numberofauthors{3} 
%
\author{
%
%
\alignauthor
    Alan J. Hamlet\\
       \affaddr{University of Florida}\\
       \affaddr{Gainesville, FL, USA}\\
       \email{AJHamlet@ufl.edu}
\alignauthor
    Carl Crane\\
       \affaddr{University of Florida}\\
       \affaddr{Gainesville, FL, USA}\\
       \email{Carl.Crane@gmail.com}
}

\maketitle
\begin{abstract}
This paper describes a novel method for allowing an autonomous ground vehicle to predict the intent of other agents in an urban environment. This method, termed the cognitive driving framework, models both the intent and the potentially false beliefs of an obstacle vehicle. By modeling the relationships between these variables as a dynamic Bayesian network, filtering can be performed to calculate the intent of the obstacle vehicle as well as its belief about the environment. This joint knowledge can be exploited to plan safer and more efficient trajectories when navigating in an urban environment. Simulation results are presented that demonstrate the ability of the proposed method to calculate the intent of obstacle vehicles as an autonomous vehicle navigates a road intersection such that preventative maneuvers can be taken to avoid imminent collisions. 
\end{abstract}

\keywords{Non-linear filtering, autonomous vehicles, prediction, multi-agent systems} 

\section{Introduction}
The potential saftey and conveinence benefits that autonomous vehicles can provide to our society are myriad. The World Health Organization reported that in 2010, 1.24 million people died due to road vehicle accidents \cite{WHO}. In addition to the potential of reducing this massive loss of life, autonomous vehicles have shown promise in increasing vehicle efficiency and convenience for drivers \cite{2,3,4}.

The vast majority of current autonomous vehicle architectures employ a reactionary response to changes in the environment. These systems require very frequent and rapid re-planning in order to avoid dynamic obstacles. Another intuitive approach is to have the autonomous vehicle predict where the dynamic obstacles are going to be in order to plan a path. One popular approach to making this prediction is to assume the dynamic obstacle continues to move in a straight line at its current velocity, as is done in the 'velocity obstacle' literature \cite{vel_obs,vel_obs2}. This approach does not take into account the control decisions made by the dynamic obstacle that affect its trajectory, as is the case for pedestrians and other vehicles.

Some research has begun to incorporate the intentions of the dynamic obstacle in order to more intelligently predict its future position. Some methods used to predict intent are hidden Markov models \cite{intent_HMM,intent_HMM2}, Markov decision processes \cite{Intent_aware_POMDP}, and Gaussian processes or mixture models \cite{hGMM,gauss_process}. These methods attempt to model trajectories and classify the dynamic obstacles' motion according to the corresponding intent. While this body of research is a step toward realizing more intelligent vehicles that truly understand their environment, it fails to consider how the obstacles' understanding of the environment will affect its future state.  

In this research, both the intent and the belief of a dynamic obstacle are considered when modeling the future states of the obstacle. This is beneficial for situations in which a dynamic obstacle, e.g. a pedestrian or another vehicle, may have an incorrect belief about the environment. For example, an obstacle vehicle trying to merge into traffic may believe it has more space than it actually does or it may not see an oncoming vehicle due to occlusions or driver error. In these situations, just knowing the driver's intent does not suffice since for the same intent she may yield or begin to merge depending on her belief. 

In this paper, the dynamics of a multiple-vehicle system are modeled as a dynamic Bayesian network. Obstacle vehicles' actions are dependent on both their intent and their belief of the surrounding environment. This idea is similar to that proposed in \cite{baker2011bayesian}, but in this paper the problem is not formulated as a Markov decision process as to avoid discretization of the state space. This is required in order to achieve the resolution necessary for the autonomous vehicle domain. Inference is performed over the network using a particle filter to jointly estimate the vehicle's intent and belief. Simulation results show that this method of joint inference allows an autonomous vehicle to predict a collision with enough time to take evasive action.

The remainder of this paper is structured as follows. In section 2, an overview of the cognitive driving framework is given. The manner of representing the system state and dynamics is described. In section 3, the process for formulating the problem as inference over a dynamic Bayesian network is explained. The structure of the DBN is detailed and the method of performing joint inference over the network using particle filtering is discussed. Next, in section 4, simulation results demonstrating the accuracy of the proposed method are presented. Finally concluding remarks are given in section 5. 

\section{The Cognitive Driving Framework}\label{sec:CDF}
This section provides an overview of the cognitive driving framework by describing how the state of an intersection environment with multiple vehicles is represented as well as the form of the system dynamics. 

In the cognitive driving framework, or CDF, the system consists of two vehicles, the obstacle vehicle and the ego vehicle, in a known environment. The joint state of the two vehicles is called the system pose and is represented as 
\begin{equation}\label{eq:system_pose}
\boldsymbol{S}_t= 
   \begin{bmatrix}
            ^{1}\boldsymbol{x}_t\\
            ^{2}\boldsymbol{x}_t\\
   \end{bmatrix}
\end{equation}
where a superscript 1 denotes the obstacle vehicle and a superscript 2 denotes the ego vehicle. In this paper, the term 'ego vehicle' refers to the vehicle running the CDF which is trying to predict the intent of the obstacle vehicle.  

In order to provide a general algorithm, the system dynamics are assumed to be nonlinear and of the form
\begin{equation}\label{eq:system_dynamics}
\boldsymbol{S}_{t+1}=
\begin{bmatrix}
^{1}\boldsymbol{x}_{t+1}\\
^{2}\boldsymbol{x}_{t+1}\\
\end{bmatrix}
=\begin{bmatrix}
f(^{1}\boldsymbol{x}_t, ^{1}\boldsymbol{u}_t,^{1}\boldsymbol{\nu}_t)\\
f(^{2}\boldsymbol{x}_t, ^{2}\boldsymbol{u}_t,^{2}\boldsymbol{\nu}_t)\\
\end{bmatrix}
\end{equation}
where \begin{math}^{i}\boldsymbol{u}_t\end{math} is the control input and \begin{math}^{i}\boldsymbol{\nu}_t\end{math} is the process noise for vehicle \emph{i} at time \emph{t}. The controller for the autonomous vehicle running the CDF (the ego vehicle) is assumed to be of the form
\begin{equation}\label{ego_control}
^{2}\boldsymbol{u}_t=h(^{2}\boldsymbol{x}_t,^{1}\boldsymbol{x}_t,^{2}I)
\end{equation} 
where the arguments to the nonlinear function \emph{h()} are the vehicle's own state, the state of the obstacle vehicle, and the intent of the ego vehicle, respectively, at time \emph{t}. The intent variable, \begin{math}
^{i}I\end{math}, represents the current behavior the vehicle is trying to execute (e.g. turn left or go straight through intersection). The control for the obstacle vehicle is modeled similarly as
\begin{equation}\label{obs_control}
^{1}\boldsymbol{u}_t=h(^{1}\boldsymbol{x}_t,\boldsymbol{B}_t,^{1}I)
\end{equation}
The difference here is that the obstacle vehicle is not assumed to have exact knowledge of the ego vehicle's state. Instead, the obstacle vehicle's controller operates on the assumed state of the ego vehicle, the \emph{belief}, \begin{math}\boldsymbol{B}_t\end{math}. It should be noted that in this context the belief is simply a point, not a distribution or density as sometimes used in the literature. If the ego vehicle has not been observed by the obstacle vehicle, then \begin{math}\boldsymbol{B}_t=\boldsymbol{\emptyset}\end{math}. The obstacle vehicle's belief updates according to the equations
\begin{equation}
\boldsymbol{B}_{t+1}=g(\boldsymbol{B}_t,\boldsymbol{O}_{t+1})
\end{equation}
\begin{equation}\label{eq:Obs}
\boldsymbol{O}_t=k(\boldsymbol{S}_t,\beta,\boldsymbol{e}_t)
\end{equation}
where \begin{math}\boldsymbol{O}_t\end{math} is the obstacle vehicle's observation at time \emph{t}, and \begin{math}\beta\end{math} is a parameter that represents the probability of the obstacle vehicle observing the ego vehicle at any given discrete time step. The observation noise, \begin{math}\boldsymbol{e}_t\end{math}, is normally distributed with a mean of zero. Given \begin{math}\boldsymbol{B}_t\end{math} and \begin{math}\boldsymbol{O}_{t+1}\end{math}, \begin{math}\boldsymbol{B}_{t+1}\end{math} updates deterministically. The observation model, \emph{k()}, determines from the system pose if the ego vehicle is in the obstacle vehicle's \emph{isovist}: the volume of space with line of sight visibility from the obstacle vehicle's pose. If the ego vehicle is occluded by other vehicles or buildings, it will not be in the obstacle vehicle's isovist, and thus \begin{math}\boldsymbol{O}_t=\boldsymbol{\emptyset}\end{math}. If the ego vehicle is in the obstacle vehicle's isovist, then the obstacle vehicle will make a noisy observation of the ego vehicle's pose with probability \begin{math}\beta\end{math}.

The goal of the cognitive driving framework is to allow the ego vehicle to predict the future states of the obstacle vehicle using this model in order to prevent collisions. This is done by performing online inference on the obstacle vehicle's belief and intent, \begin{math}\boldsymbol{B}_t\end{math} and \begin{math}I_t\end{math}. The following section details the procedure for performing this joint inference.

\begin{figure}
\includegraphics[scale=0.5]{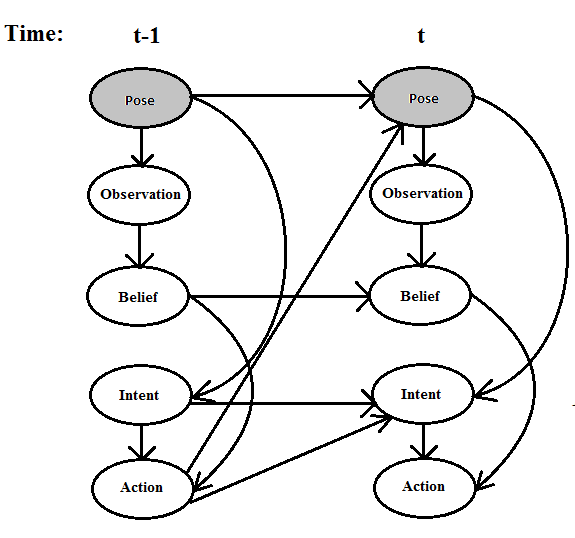} \caption{The structure of the DBN used in the cognitive driving framework.}\label{fig:DBN}
\end{figure}

\section{Joint Belief-Intent Inference}
In this section, the model outlined in the previous section is formulated as a dynamic Bayesian network, or DBN. The relationships between the variables in the network are described and how inference is performed over the network using a particle filter is explained.

\subsection{Dynamic Bayesian Network}
The cognitive driving framework uses a dynamic Bayesian network to capture the dependencies between the random variables in the CDF system dynamics. A Bayesian network is a directed acyclic probabilistic graphical model that is used to represent a set of random variables and their conditional dependencies. A \emph{dynamic} Bayesian network is a Bayesian network which relates the variables to each other over sequential time steps. Sometimes dynamic Bayesian networks are called \emph{two-time-slice Bayesian networks} because at any point in time \emph{t}, the value of a variable in the network can be calculated from the prior value (at time \emph{t-1}) and the independent variables. Kalman filter models and hidden Markov models are special cases of DBN's, but, in general, DBN's allow the hidden state of the system to be factored into separate variables so the structure of the dependencies between the variables can be exploited.

The structure of the DBN used in the CDF is depicted in figure \ref*{fig:DBN}. This graphical model reflects the dependencies given by the equations in section \ref*{sec:CDF}. The gray nodes in the graph denote the variable known by the ego vehicle, the system pose, \begin{math}\boldsymbol{S}_t\end{math}, as given in equation \ref*{eq:system_pose}. In some contexts, because the value of this variable is provided to the ego vehicle by its sensors, it is called the observation. In this work the observation, \begin{math}
\boldsymbol{O}_t\end{math}, refers to the obstacle vehicle's noisy measurement of the ego vehicle's pose, \begin{math}
^{2}\boldsymbol{x}_t\end{math}.

Between time-slices, the variables in the DBN flow temporally from left to right and within a time-slice they flow (more-or-less) from top to bottom. The system pose affects the obstacle vehicle's observation which in turn determines the obstacle vehicle's belief. The obstacle vehicle's intent and belief of the system pose inform the obstacle vehicle's action. The joint actions of the two vehicles result stochastically in the next system pose. Without loss of generality, the intent of the obstacle vehicle is assumed to be constant throughout an episode. 

\subsection{Filtering}

\begin{figure}
\includegraphics[scale=0.4]{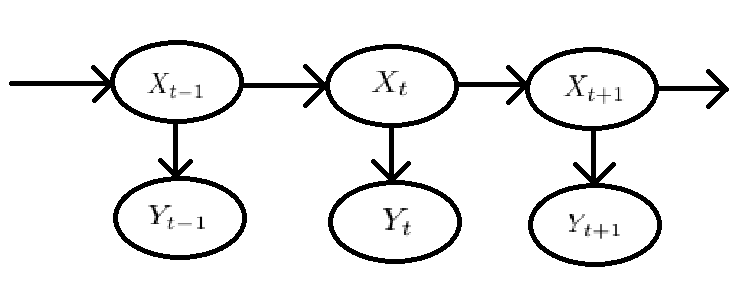} \caption{Simplified model of the CDF using the joint DBN state variable \protect\begin{math}X_t\protect\end{math}.} \label{fig:HMM}
\end{figure} 

Now that the two-vehicle system dynamics are represented as a DBN, a method of filtering needs to be implemented in order to perform online inference of the obstacle vehicle's belief and intent. By combining equations \ref*{eq:system_dynamics}-\ref*{eq:Obs} we can represent the DBN state and dynamics, repectively, as \begin{equation}
\boldsymbol{X}_t=[\boldsymbol{S}_t,\boldsymbol{B}_t,I]^T
\end{equation}
\begin{equation}
\boldsymbol{X}_{t+1}=F(\boldsymbol{X}_t)      
\end{equation}
This condenses the DBN in figure \ref*{fig:DBN} to that shown in figure \ref*{fig:HMM}, which is the typical representation for filtering problems. The variable \begin{math}\boldsymbol{Y}_t\end{math} represents the measurement, which in this study is the system pose, \begin{math}\boldsymbol{S}_t\end{math}. 

As shown in section \ref*{sec:CDF}, the sytem dynamics are highly non-linear. Instead of linearizing the dynamics at the expense of accuracy of the estimation, a non-linear Monte-Carlo based filtering method was employed. Monte-Carlo (MC) methods are ideally suited for the current application due to their ability to model highly non-linear systems with multi-modal, non-Gaussian distributions. In this study, the DBN state is composed of both continuous and discrete variables, representing discrete intention hypotheses making traditional linearization methods such as the Extended or Unscented Kalman Filter unsuitable for this application.

Particle filters are sequential Monte-Carlo methods that maintain an estimate of the posterior distribution of the system state as a set of particles. This non-parametric representation is capable of representing arbitrarily complex distributions as long as a large enough particle set is used. Each particle is initialized according to the a priori distribution and is propagated through the noisy system dynamics. The particles are then resampled according to the particles' importance weights. The importance weight of a particle is proportional to the likelihood of the particle generating the measurement, \begin{math}\boldsymbol{Y}_t=\boldsymbol{S}_t\end{math}. In this study, the likelihood is represented as a Gaussian distribution centered around the measured system pose. The weight of each particle is proportional to the probability of the system pose of the particle given the measured system pose, as shown in the equation below.
\begin{equation}
w_t^{[m]}\propto P(\boldsymbol{X}_t^{[m]}|\boldsymbol{S}_t) \sim{} \protect\mathcal{N}(\boldsymbol{S}_t,\boldsymbol{\Sigma})
\end{equation}
\begin{math}\boldsymbol{\Sigma}\end{math} is the variance of the Gaussian likelihood function. A superscript \begin{math}[m]\end{math} denotes that the variable corresponds to the \begin{math}m^{th}\end{math} particle. The weights are normalized such that they sum to one.

Once the importance weights are calculated, the particles are resampled in order to move the posterior distribution toward the region of the state space that matches the measurement. The technique of \emph{stratified} or \emph{low-variance} sampling is employed to reduce computational complexity \cite{StratifiedSampling}. The resulting set of particles is an approximation to the actual state of the system. As the number of particles, \emph{N}, approaches infinity, the particle set converges to the true distribution of the state. 

\section{Experimental Results}
In order to demonstrate the ability of the CDF to perform joint inference on the intent and belief of an obstacle vehicle, a simulated experiment is conducted. This section describes the simulation set up and presents the results showing the CDF's ability to detect and prevent vehicle collisions in a road intersection navigation scenario.

\subsection{Simulation}
\begin{figure}
\centering
\includegraphics[scale=0.4]{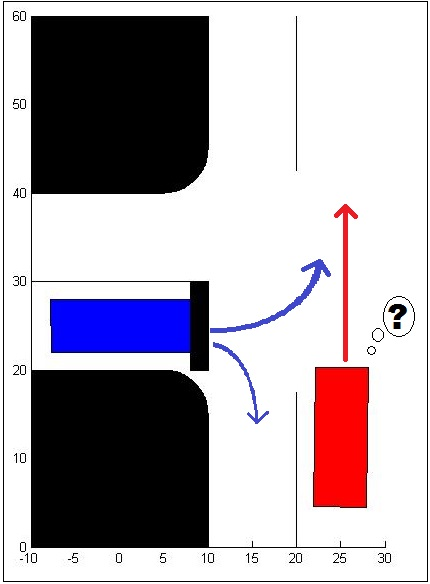} \caption{A close up image of the simulated intersection environment with the ego vehicle (red) trying to predict the intent of the obstacle vehicle (blue).} \label{fig:Intersection}
\end{figure}

The CDF was tested using a simulated T-intersection scenario as depicted in figure \ref*{fig:Intersection}. The autonomous vehicle running the CDF (the ego vehicle, in red) has the right of way. An obstacle vehicle (blue) is stopped at a stop sign at the T-intersection and can either turn left into the ego vehicle's lane or turn right. It is desirable for the ego vehicle to predict not just the intention of the obstacle vehicle to turn left, but whether the obstacle vehicle is going to turn left in front of the ego vehicle or if it is going to yield. 

The simulation uses a bicycle kinematic model for the vehicles as described in \cite{hGMM}. The pose for a vehicle from equation \ref*{eq:system_pose} is given by the four dimensional vector
\begin{equation}
\boldsymbol{x}_t=
	\begin{bmatrix}
           x_t\\
           y_t\\
           \theta_t\\
           v_t\\
   \end{bmatrix}
\end{equation}
where \begin{math}x_t\end{math} and \begin{math}y_t\end{math} are the position of the rear axle in the ground plane, \begin{math}
\theta_t\end{math} is the vehicle's orientation, and \begin{math}v_t\end{math} is the speed of the vehicle, all at time \emph{t}. The superscript indicating which vehicle the state corresponds to has been left off here for clarity. The dynamics from equation \ref{eq:system_dynamics} are then given by
\begin{equation}
\boldsymbol{x}_{t+1}=f(\boldsymbol{x}_t,\boldsymbol{u}_t,\boldsymbol{\nu}_t)=\begin{bmatrix}
            x_t+ v_t \cdot \Delta t \cdot \cos{\theta_t}\\
            y_t+ v_t \cdot\Delta t \cdot \sin{\theta_t}\\
            \theta_t+\frac{v_t \cdot \Delta t}{l} \cdot \tan{(_{2}u_t+_{2}\nu_t)}\\
            v_t+(_{1}u_t+_{1}\nu_t) \cdot \Delta t\\
   \end{bmatrix}
\end{equation}
where the elements of the two dimensional control input are acceleration, \begin{math}_{1}u_t\end{math}, and steering angle, \begin{math}
_{2}u_t\end{math}. The two components of the process noise, \begin{math}_{1}\nu_t\end{math} and \begin{math}_{2}\nu_t\end{math}, are both zero mean Gaussian noise affecting the realization of the controller's commanded acceleration and steering angle. The parameter \begin{math}l\end{math} is the wheelbase of the vehicle. In these simulations, a time step, \begin{math}\Delta t\end{math}, of 0.1 seconds is used.

The controller used in this simulation has a different path following controller for each intent, \emph{I}, that is coupled with a hand tuned finite state machine that determines whether the vehicle should yield to the other vehicle or if it is clear to proceed. As shown in equations \ref*{ego_control} and \ref*{obs_control}, the ego vehicle determines its control input based on the known positions of both vehicles, while the obstacle vehicle only has access to its own position and a noisy estimate of the ego vehicle's position.

\subsection{Results and Analysis}
Experiments were performed to compare the CDF to a purely reactionary planner. The simulation described in the previous section was run with the obstacle vehicle randomly choosing to turn left with a probability of 0.75 or right with a probability of 0.25. The parameter \begin{math}\beta\end{math} in equation \ref*{eq:Obs} was set to 0.05. This selection of \begin{math}\beta\end{math} corresponds to about a 40 percent chance of the obstacle vehicle observing the ego vehicle within the first second of simulation. The ego vehicle has the right-of-way and is attempting to drive through the intersection at a speed of about 30 miles per hour (48 \begin{math}km/h)\end{math}.

During the first experiment, the ego vehicle is using the CDF to jointly estimate the obstacle vehicle's belief and intent. If the obstacle vehicle's predicted intent is to turn left and its belief is that the intersection is clear, then a collision is predicted to be imminent. The ego vehicle then brakes at the maximum rate in an attempt to avoid the collision. The maximum rate of deceleration used in the experiments was 16 \begin{math}ft/s^2\end{math}, which is reasonable for a passenger vehicle traveling on a road surface with a moderate coefficient of friction \cite{braking}. 

\begin{figure}
\begin{center}\begin{tabular}{| l | c | c | c || r |}
\hline
 & \multicolumn{3}{|c||}{\bf{Prediction}} & \\ \cline{2-5}
  \bf{Scenario} & Cutoff & Yield & Right & Total \\ \hline
  Cutoff & 417 & 0 & 0 & 417\\ \hline
  Yield & 13 & 334 & 0 & 347\\ \hline
  Right & 0 & 0 & 236 & 236\\ \hline
  \multicolumn{4}{c|}{ } & 1000\\ \cline{5-5}
\end{tabular}\end{center}
\caption{Simulation results using the cognitive driving framework. Each row corresponds to the scenario given in the leftmost column while the columns the prediction given by the CDF.} \label{CDF_results}
\end{figure}

\begin{figure}
\begin{center}\begin{tabular}{| l | c | c | r |}
\hline
   & Collisions & Collisions & Collisions \\
   & Imminent & Occured & Avoided \\ \hline
  CDF & 417 & 9 & 98\% \\ \hline
  Reactive & 404 & 213 & 47\% \\ \hline
\end{tabular}\end{center}
\caption{Comparison of simulation results for the CDF and a purely reactionary planner. } \label{results}
\end{figure}

The table in figure \ref*{CDF_results} details the simulation results. The simulation was run for a total of 1000 episodes and the obstacle vehicle cutoff the ego vehicle a total of 417 times. In this context, 'cutoff' means that if the ego vehicle were to keep its speed constant and not take preventative measures, a collision would result. While 13 scenarios where falsely classified as imminent collisions, 100\% of the true imminent collisions were recognized by the CDF. These false positives due to the obstacle vehicle observing the ego vehicle and braking at the last second. 

In the next experiment, the same simulation was run except a reactionary planner was used in place of the CDF. If any part of the obstacle vehicle entered the ego vehicle's lane, the ego vehicle performed an emergency braking maneuver. The planner was run at a rate of 10 Hz. After a total of 1000 episodes, the ego vehicle was cutoff 404 times. The table in figure \ref*{results} compares the performance of the reactive planner to that of the CDF. The CDF was able to avoid 98\% of the imminent collisions caused by the obstacle vehicle, while the reactive planner was only able to avoid 47\% of the imminent collisions.

\section{Conclusions}
This paper presented the cognitive driving framework, a method for joint inference of the intent and belief of an obstacle vehicle in an intersection navigation scenario. The goal of the CDF is to allow an autonomous vehicle to predict when a potentially hazardous situation is about to occur early enough to allow the autonomous vehicle to take evasive action to prevent a collision. The formulation of the problem as a dynamic Bayesian network was presented. A non-linear filtering method was proposed using a particle filter to estimate the posterior distribution of the state of the DBN. Finally, the accuracy of the estimation method was demonstrated by simulating an intersection navigation scenario where an obstacle vehicle cuts off the autonomous vehicle. The simulation results show that the CDF is able to predict and prevent 98\% of imminent collisions. For comparison, the same simulation was run using a purely reactionary planner, resulting in only 47\% of imminent collisions being avoided. 


%
\bibliographystyle{abbrv}
\bibliography{fcrarproc}  
\balancecolumns
\end{document}